\documentclass[10pt,conference]{IEEEtran}
\IEEEoverridecommandlockouts
\usepackage{cite}
\usepackage{amsmath,amssymb,amsfonts}
\usepackage{algorithmic}
\usepackage{graphicx}
\usepackage{textcomp}
\usepackage{xcolor}
\usepackage{booktabs} 
\usepackage{multirow}
\usepackage{bm}
\usepackage{float}
\usepackage{placeins}
\usepackage{caption}
\usepackage{xcolor}
\usepackage{mathptmx} 
\usepackage{times}
\usepackage{placeins}
\def\BibTeX{{\rm B\kern-.05em{\sc i\kern-.025em b}\kern-.08em
    T\kern-.1667em\lower.7ex\hbox{E}\kern-.125emX}}
\begin{document}

\title{G-ICSO-NAS: Shifting Gears between Gradient and Swarm for Robust Neural Architecture Search\\

}

\author{
  \IEEEauthorblockN{
    1\textsuperscript{st} Xingbang Du\textsuperscript{\dag},~
    2\textsuperscript{nd} Enzhi Zhang\textsuperscript{\ddag},~
    3\textsuperscript{rd} Rui Zhong\textsuperscript{\ddag},~
    4\textsuperscript{th} Yang Cao\textsuperscript{\dag},~
    5\textsuperscript{th} Masaharu Munetomo\textsuperscript{\ddag}
  }
  \IEEEauthorblockA{
    \textsuperscript{\dag}\textit{Information Systems Design Laboratory,
    Hokkaido University}, Sapporo, Japan\\
    \textsuperscript{\ddag}\textit{Information Initiative Center,
    Hokkaido University}, Sapporo, Japan\\
    xingbang.du.d1@elms.hokudai.ac.jp,~
    \{zhangenzhi, zhongrui, munetomo\}@iic.hokudai.ac.jp,~
    yang.cao.y4@elms.hokudai.ac.jp
  }
}

\maketitle

\begingroup
\renewcommand\thefootnote{}
\footnotetext{Accepted at the International Joint Conference on Neural Networks (IJCNN), WCCI 2026. This is a preprint version.}
\endgroup

\begin{abstract}
Neural Architecture Search (NAS) has become a pivotal technique in automated machine learning. Evolutionary Algorithm (EA)-based methods demonstrate superior search quality but suffer from prohibitive computational costs, while gradient-based approaches like DARTS offer high efficiency but are prone to premature convergence and performance collapse. To bridge this gap, we propose G-ICSO-NAS, a hybrid framework implementing a three-stage optimization strategy. The Warm-up Phase pre-trains supernet weights ($w$) via differentiable methods while architecture parameters ($\alpha$) remain frozen. The Exploration Phase adopts a hybrid co-optimization mechanism: an Improved Competitive Swarm Optimizer (ICSO) with diversity-aware fitness navigates the architecture space to update $\alpha$, while gradient descent concurrently updates $w$. The Stability Phase employs fine-grained gradient-based search with early stopping to converge to the optimal architecture. By synergizing ICSO's global navigation capability with differentiable methods' efficiency, G-ICSO-NAS achieves remarkable performance with minimal cost. In the context of the DARTS search space, an accuracy of 97.46\% is achieved on CIFAR-10 with a computational budget of just 0.15 GPU-Days. The method also exhibits strong transfer potential, recording accuracies of 83.1\% (CIFAR-100) and 75.02\% (ImageNet). Furthermore, regarding the NAS-Bench-201 benchmark, G-ICSO-NAS is shown to deliver state-of-the-art results across all evaluated datasets.
\end{abstract}

\begin{IEEEkeywords}
Neural Architecture Search, Competitive Swarm Optimizer, Differentiable Architecture Search
\end{IEEEkeywords}

\section{Introduction}
Deep Convolutional Neural Networks (CNNs) have emerged as the predominant approach for image processing, demonstrating exceptional efficacy in various computer vision applications, largely attributed to the evolution of landmark architectures represented by VGG \cite{simonyan2014very}, ResNet \cite{he2016deep}, as well as MobileNet \cite{howard2017mobilenets} and EfficientNet \cite{tan2019efficientnet}, which have established performance benchmarks for visual representation learning. Nevertheless, the conventional hand-crafting of neural architectures necessitates significant domain expertise and iterative testing, making the procedure notoriously time-consuming and resource-demanding.

Consequently, Neural Architecture Search (NAS) has emerged as an automated paradigm to alleviate the substantial burden of manual architecture design. While initial NAS strategies based on Evolutionary Algorithms (EAs) and Reinforcement Learning (RL) attained superior performance, their feasibility was severely compromised by an overwhelming computational burden. \cite{pham2018efficient}. To address this efficiency bottleneck, Differentiable Architecture Search (DARTS) \cite{liu2018darts} was proposed, which relaxes the discrete search space into a continuous one, enabling efficient gradient-based optimization and dramatically accelerating the search process. Nevertheless, DARTS suffers from critical robustness challenges in its supernet training. The most prominent issue is the so-called "collapse" phenomenon \cite{zela2019understanding}, where the search process exhibits a strong bias toward parameter-free operations (e.g., none or skip\_connect) in early stages, leading to degraded final performance. Additionally, the inherent instability of the bilevel optimization process often results in suboptimal discrete architectures.

In contrast, EAs demonstrate complementary strengths to gradient-based differentiable methods in the NAS domain. While evolutionary approaches offer superior exploration capabilities and robustness to local optima at the cost of computational efficiency, differentiable methods excel in search speed through gradient-based optimization but suffer from the aforementioned stability issues \cite{elsken2019neural}. This fundamental complementarity in their computational efficiency and architecture quality trade-offs provides a strong theoretical foundation for their synergistic integration.

This study proposed G-ICSO-NAS, a novel three-stage method that navigates the NAS landscape by strategically shifting gears between gradient-driven descent and swarm-based exploration, striking a robust balance between computational efficiency and architecture quality. Our core innovations are as follows:

We designed Double-helix parameter update mechanism to achieve high quality of solution: During the exploration stage, architecture parameters $\alpha$ are updated by the ICSO algorithm through swarm-based evaluation (requiring only forward propagation), while super-net weights w are updated through standard epoch-based training. While DARTS' simultaneous gradient updates on weights and architecture parameters cause rapid convergence along the path of least resistance to local optima, ICSO's population-based evaluation of multiple $\alpha$ configurations preserves diversity.

Multi-dimensional fitness function is designed to avoid local optimal: The ICSO algorithm in the exploration stage incorporates three types of fitness functions: a validation-accuracy-based function selects the best individual to guide the update of architecture parameters $\alpha$, while swarm diversity and operation diversity functions, combined with loss-based fitness, jointly guide the swarm evolution direction.The validation-loss-based fitness and the two diversity-based fitness functions serve complementary yet synergistic roles: the former ensures consistent progression toward high-accuracy architectures, while the latter mitigate the risk of entrapment in local optima—an advantage unattainable when relying solely on validation loss as the optimization objective.

Adaptive termination strategy aims to reduce computational costs: In the stabilization stage, we employ the DARTS method with a reduced architecture update learning rate and implement a convergence testing mechanism for adaptive search termination. 

Finally, the proposed method has been extensively validated across different datasets and search spaces: On CIFAR-10, the search achieves 97.46\% accuracy with only 0.15 GPU-days. The discovered architecture demonstrates excellent transferability, attaining 83.10\% accuracy on CIFAR-100 and 75.02\% (Top-1) / 92.51\% (Top-5) accuracy on ImageNet. Specifically within the search space of NAS-Bench-201, our method identifies the architecture with the highest test accuracy, surpassing all compared state-of-the-art methods.

\section{Related Work}
\subsection{Evolution Algorithm for NAS}
NAS marks a milestone in deep learning's transition from manual tuning to automated design. This field was inaugurated by reinforcement learning (RL)-based methods from Google Brain, demonstrating machines' capability to discover state-of-the-art architectures within vast search spaces. To improve efficiency, researchers subsequently introduced cell-based search spaces \cite{zoph2018learning}.

Concurrently, evolutionary algorithm (EA)-based approaches achieved remarkable results. Through mutation and tournament selection, EAs exhibit strong global search robustness, successfully yielding AmoebaNet—an architecture surpassing contemporary hand-designed models. These studies demonstrated EAs' potential \cite{real2017large} for discovering non-intuitive architectures.
Although first-generation methods achieved peak performance, their prohibitive computational costs—often thousands of GPU hours—motivated more efficient paradigms, paving the way for parameter sharing techniques like ENAS \cite{pham2018efficient} and differentiable search methods.

\subsection{Differentiable architecture search}
DARTS \cite{liu2018darts} pioneered the differentiable search paradigm by projecting discrete spaces into a continuous domain. This relaxation enables gradient-based updates, where the search efficiency is achieved by framing the learning process as a bilevel optimization problem that alternates between architecture parameters and model weights.Specifically, the output of each edge in the search cell is computed as a softmax-weighted combination of candidate operations:
\begin{equation}
    \bar{o}^{(i,j)}(x) = \sum_{o \in \mathcal{O}} \frac{\exp(\alpha_o^{(i,j)})}{\sum_{o' \in \mathcal{O}} \exp(\alpha_{o'}^{(i,j)})} o(x)
\end{equation}
where $\mathcal{O}$ denotes the set of candidate operations, and $\alpha$ represents the architecture parameters. DARTS alternately optimizes the network weights $w$ and architecture parameters $\alpha$ via gradient descent, enabling efficient end-to-end search.

Subsequent variants addressed its limitations: PC-DARTS \cite{xu2019pc} improved memory efficiency through partial channel sampling, enabling search on larger-scale datasets; P-DARTS \cite{chen2019progressive} bridged the depth gap between search and evaluation via progressive depth increase and search space regularization; R-DARTS \cite{zela2019understanding} analyzed failure mechanisms of the original algorithm and proposed Hessian eigenvalue monitoring to prevent performance degradation; and FBNet \cite{wu2019fbnet} incorporated hardware-aware latency prediction, enabling joint optimization of accuracy and inference speed for resource-constrained deployment.

\subsection{Competitive Swarm Optimizer}
To tackle high-dimensional optimization problems, the Competitive Swarm Optimizer (CSO) \cite{cheng2014competitive} was introduced as a specialized PSO variant. It eliminates the reliance on global best tracking in favor of a pairwise competition mechanism, where losing particles update their positions based on the winners. Instead, particles are randomly paired for competition: the loser learns from the winner, while the winner passes directly to the next generation. The loser particle updates its velocity and position as follows:
\begin{equation}
    v_l(t+1) = R_1 \cdot v_l(t) + R_2 \cdot (x_w(t) - x_l(t)) + \phi R_3 \cdot (\bar{x}(t) - x_l(t))
\end{equation}
where $x_w$ and $x_l$ denote the winner and loser positions respectively, $\bar{x}$ is the swarm mean position, $R_1, R_2, R_3$ are random vectors in $[0,1]$, and $\phi$ controls the influence of the swarm centroid.

To further enhance search efficiency, several notable extensions have been proposed: SL-PSO \cite{cheng2015social} introduces a social learning mechanism allowing particles to learn from any better demonstrators to improve population diversity; and LLSO \cite{yang2017level} employs a level-based structure where particles learn from superior hierarchy levels to better balance exploration and exploitation. These variants have demonstrated robust performance on high-dimensional benchmarks, establishing the learning-based swarm framework as a powerful alternative to classical PSO.

\section{Methodology} 
In this section,  we will first present a comprehensive overview of the proposed G-ICSO-NAS method. Subsequently, the detailed designs of the Warmup Stage, Exploration Stage, and Stability Stage are elaborated in sub-sections B, C, and D, respectively.

\begin{figure*}[htbp]
	\includegraphics[width=\textwidth]{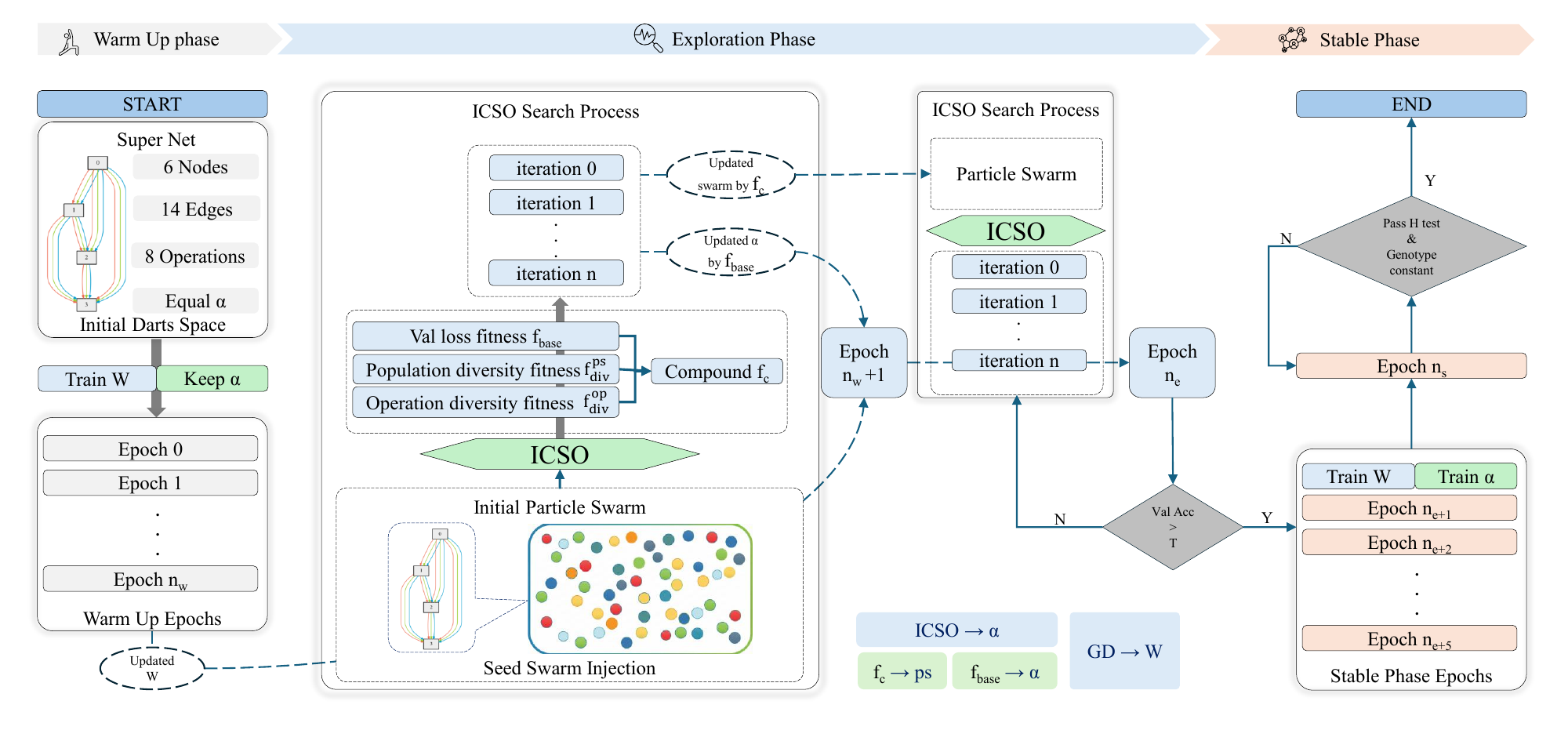}
	\caption{The overview of G-ICSO-NAS. The warm-up stage trains super-net weights $W$ with unchanged architecture weights $\alpha$; the exploration stage alternates between ICSO-based  $\alpha$ updates and gradient-based $W$ updates; the stability stage performs gradient-based optimization with Hoeffding-based early stopping for convergence detection.}
	\label{fig1}
\end{figure*}

\subsection{Overview Of G-ICSO-NAS}

Figure~\ref{fig1} depicts the overall workflow of the G-ICSO-NAS method. It comprises three distinct phases: the Warmup Stage, the Exploration Stage, and the Stability Stage. In the Warmup Stage, we adhere to the supernet design principles of DARTS to train and update the supernet weights $w$, while the architecture parameters $\alpha$ remain frozen to ensure a stable initialization. The subsequent Exploration Stage synergizes ICSO with supernet training; specifically, the former iteratively searches for optimal combinations of architecture parameters $\alpha$, while the latter continues to update the super-net weights $w$ via gradient descent. Finally, in the Stability Stage, the method reverts to a gradient-based DARTS approach. Building upon the architecture parameters discovered during the exploration phase, we fine-tune $\alpha$ using a reduced learning rate and terminate the search process upon satisfying the Hoeffding Inequality-based convergence test.

\subsection{Warm-Up Stage}

In this subsection, we elaborate on the first phase: the Warm-up Stage. As illustrated in the left panel of Fig.~\ref{fig1}, a supernet is constructed based on the DARTS search space to serve as the initialization for architecture search. Each intermediate node $j$ in the cell aggregates information from all predecessor nodes:
\begin{equation}
\label{eq:supernet_node}
x^{(j)} = \sum_{i<j} \bar{o}^{(i,j)}(x^{(i)})
\end{equation}
where the mixed operation $\bar{o}^{(i,j)}$ is defined as a softmax-weighted sum over all candidate operations:
\begin{equation}
\label{eq:mixed_op}
\bar{o}^{(i,j)}(x) = \sum_{o \in \mathcal{O}} \frac{\exp(\alpha_o^{(i,j)})}{\sum_{o' \in \mathcal{O}} \exp(\alpha_{o'}^{(i,j)})} o(x)
\end{equation}
where $\mathcal{O}$ denotes the set of candidate operations, and $\alpha_o^{(i,j)}$ represents the architecture weight for operation $o$ on edge $(i,j)$.

Subsequently, the supernet is trained to update the network weights $w$ using training data, while the architecture parameters $\bm{\alpha}$ remain frozen (by setting the architecture learning rate to zero), ensuring equal probability for all operations:
\begin{equation}
\label{eq:warmup_update}
\begin{aligned}
w_{t+1} &\leftarrow w_t - \eta_w \nabla_w \mathcal{L}_{\text{train}}(w_t, \bm{\alpha}_{\text{init}}) \\
\bm{\alpha}_{t+1} &\leftarrow \bm{\alpha}_{t} = \bm{\alpha}_{\text{init}}
\end{aligned}
\end{equation}

Upon completion of the predefined warmup epochs, the warmup phase terminates. At this point, the supernet weights $w$ have stabilized, preparing the method for the subsequent exploration phase.

\subsection{Exploration Stage}

In this subsection, we detail the second phase of the method: the Exploration Stage. 

We begin by introducing the Improved Competitive Swarm Optimizer (ICSO), which is a simplified variant of L-ICSO \cite{zhong2025improved}, Improved Competitive Swarm Optimizer with Linear Population Reduction, originally proposed by Zhong et al. in 2025. The core advantage of the original L-ICSO lies in its integration of a novel triplet individual competition mechanism to enhance optimization performance, coupled with a linear swarm size reduction strategy. L-ICSO was pitted against ten advanced competitors across the CEC2017 benchmark functions. Analysis of the experimental results demonstrates that L-ICSO possesses significant advantages in solving high-dimensional problems, validating both its convergence speed and solution quality through statistical tests.

Regarding the search within the DARTS space, due to the continuous relaxation of the architecture representation (where each edge is parameterized by a vector of operation weights), this search task is mathematically equivalent to a 224-dimensional optimization problem:

\begin{equation}
\label{eq:dimension_calc}
D = 2 \times K \times |\mathcal{O}| = 2 \times \left( \sum_{i=0}^{N-1} (i+2) \right) \times |\mathcal{O}|
\end{equation}
where $K$ denotes the number of edges per cell, $N$ is the number of intermediate nodes, and $|\mathcal{O}|$ is the number of candidate operations.

The architecture search is formulated as:
\begin{equation}
\label{eq:optimization_problem}
\min_{\bm{x} \in \mathbb{R}^D} \quad f(\bm{x}) = \mathcal{L}_{\text{val}}(w^*, \bm{x})
\end{equation}
where $\bm{x} = [\text{vec}(\bm{\alpha}_{\text{normal}}), \text{vec}(\bm{\alpha}_{\text{reduce}})]$ is the concatenated architecture parameter vector.

 Although the evaluation of a candidate architecture primarily requires forward propagation without gradient backpropagation, assessing a substantial swarm of architectures remains computationally intensive. Therefore, the L-ICSO algorithm, distinguished by its optimal balance of efficiency and effectiveness, is exceptionally well-suited as the search engine for this exploration phase.

Secondly, to preserve architectural diversity, we discard the swarm reduction mechanism and introduce two novel diversity-aware fitness functions into ICSO. Specifically, the swarm Diversity Function is designed to maintain individual diversity throughout the iterative process:

\begin{equation}
f_{div}^{swarm}(\vec{x}) = \tanh\left( \min_{\bm{x}' \in \mathcal{H}} \frac{\|\bm{x} - \bm{x}'\|_2}{\sqrt{D}} \right)
\end{equation}
where $\mathcal{H}$ is the historical swarm set and $D$ is the parameter dimension.

The Operation Diversity Function is formulated to mitigate the collapse of the search space, suppressing the dominance of parameter-free operations (e.g., none and skip\_connect) often observed in the early stages of DARTS:

\begin{equation}
f_{div}^{op}(\vec{x}) = \frac{-\sum_{o \in \mathcal{O}} p_o \log p_o}{\log |\mathcal{O}|}
\end{equation}
where $p_o$ is the selection frequency of operation $o$ in the architecture, and $|\mathcal{O}|$ is the number of candidate operations.

By integrating these two diversity metrics with the Basic Fitness Function which derived from the architecture's forward propagation loss,

\begin{equation}
f_{\text{base}}(\bm{x}) = \frac{\mathcal{L}_{\text{val}}(\bm{x}) - L_{\min}}{L_{\max} - L_{\min}}
\end{equation}

where $\mathcal{L}_{\text{val}}(\bm{x})$ is the cross-entropy loss on the validation set, $L_{\min}$ and $L_{\max}$ are the expected loss bounds. Lower fitness indicates better architecture.

We propose a Combined Fitness Function for swarm updates:

\begin{equation}
f_{c}(\vec{x}) = f_{\text{base}}(\bm{x}) - \lambda_{\text{swarm}} \cdot f_{div}^{swarm}(\vec{x}) - \lambda_{\text{op}} \cdot f_{div}^{op}(\vec{x})
\end{equation}

where $\lambda_{\text{swarm}}$ and $\lambda_{\text{op}}$ are the weights for particle swarm diversity and operation diversity, respectively. The subtraction encourages higher diversity since ICSO minimizes the fitness function.

Finally, as depicted in the middle section of Fig.~\ref{fig1}, the Basic Fitness is utilized to select the optimal $\alpha$ parameters of the current iteration to guide the architecture update for the subsequent epoch, whereas the Combined Fitness is employed to drive the evolution of the ICSO swarm, thereby ensuring sustained diversity.

The workflow of the refined ICSO algorithm is illustrated in Fig.~\ref{fig2}. Initially, the fitness of the swarm is evaluated using the Combined Fitness Function, and the global optimum solution is recorded. Subsequently, the swarm indices are randomly shuffled, and the particles are partitioned into triplets. Within each triplet, the fitness values of the three particles are compared to categorize them into three hierarchical tiers: the Winner ($x_w$), the Second-Best ($x_m$), and the Loser ($x_l$). Finally, the Winner is directly preserved for the next generation without updating its architecture parameters $\alpha$. The Second-Best particle has a 50\%  probability of remaining unchanged (direct replication) and a 50\% probability of being updated. In the update case, it primarily learns from the triplet's Winner while adjusting based on the swarm's centroid ($x_{mean}$):

\begin{figure}[htbp]
	\includegraphics[width=0.5\textwidth]{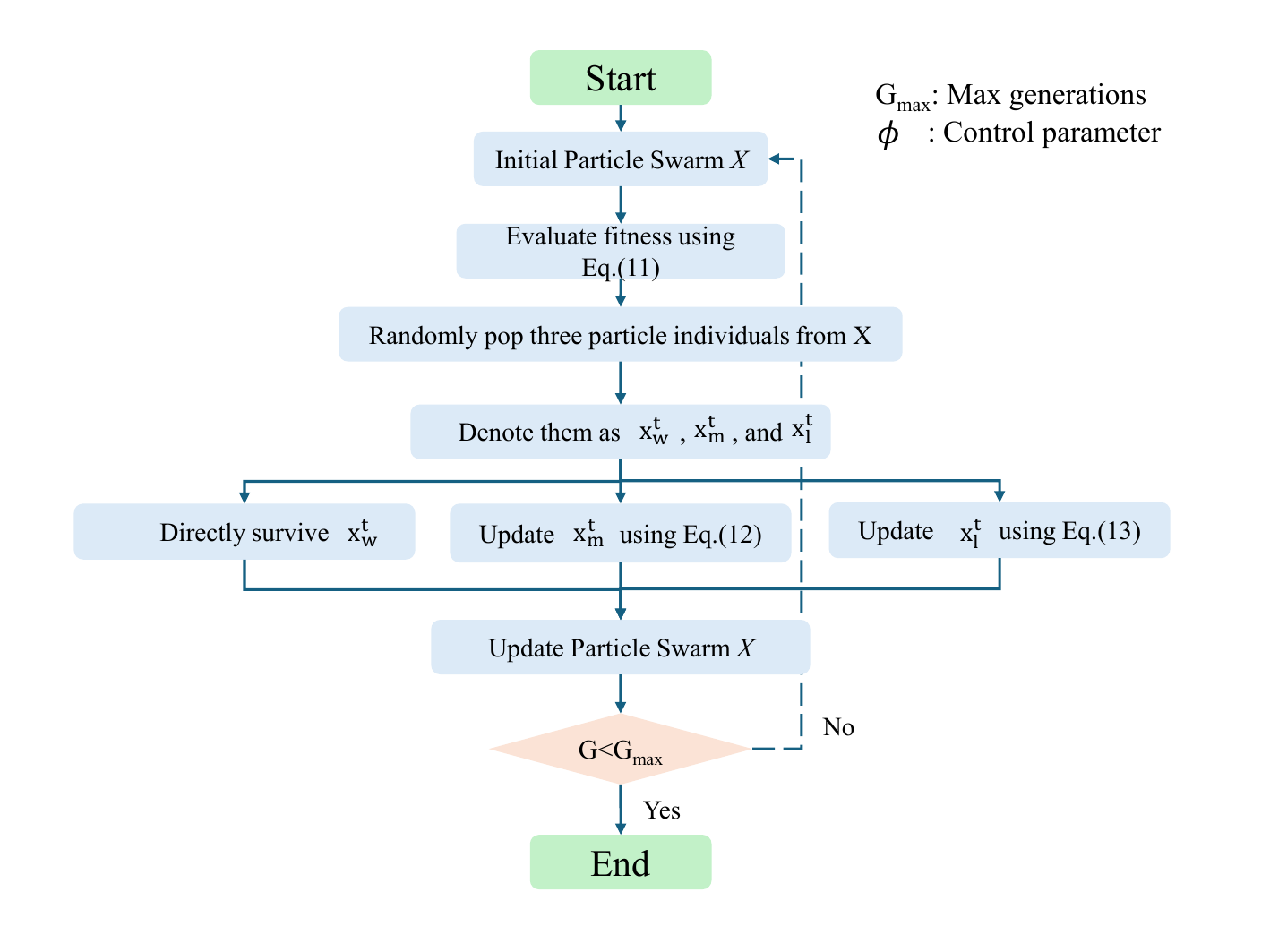}
	\caption{ICSO. a triplet-based competitive swarm optimizer with diversity-aware fitness evaluation.}
	\label{fig2}
\end{figure}

\begin{equation}
\label{eq:icso_second_best}
\begin{aligned}
&\text{If } \text{rand}(0, 1) < 0.5: \\
&\quad v_m^{t+1} = r_1 v_m^t + r_2 \cdot (x_w^t - x_m^t) + \phi \cdot r_3 \cdot (x_{mean}^t - x_m^t) \\
&\quad x_m^{t+1} = x_m^t + v_m^{t+1} \\
&\text{Else:} \\
&\quad x_m^{t+1} = x_m^t
\end{aligned}
\end{equation}

Conversely, the Loser is mandatorily updated. Its trajectory is adjusted by learning from the triplet's Winner and incorporating information from the global best individual ($x_{best}$):

\begin{equation}
\label{eq:icso_loser}
\begin{aligned}
v_l^{t+1} &= r_1 v_l^t + r_2 \cdot (x_w^t - x_l^t) + \phi \cdot r_3 \cdot (x_{best}^t - x_l^t) \\
x_l^{t+1} &= x_l^t + v_l^{t+1}
\end{aligned}
\end{equation}

where $\phi$ is the control parameter that regulates the learning strength of the second-best or loser particle during the competitive update, and $r_1, r_2, r_3$ denote dimension-wise random vectors which introduce stochasticity into the update process.

Finally, as depicted in the middle section of Figure~\ref{fig1}, the workflow of the Exploration Stage proceeds as follows. First, an initial population is randomly generated within the search space:
\begin{equation}
\label{eq:pop_init}
\mathcal{P}_0 = \{\bm{x}_i \mid \bm{x}_i \sim U(\bm{l}, \bm{u}), \ i = 1, \dots, N_{\text{pop}}\}
\end{equation}
where $\bm{l}$ and $\bm{u}$ denote the lower and upper bounds of the search space. While all individuals within this population possess distinct architecture parameters, they share the supernet weights $w$ inherited from the Warmup Stage.

Subsequently, ICSO is employed to evolve the population based on the combined fitness function, yielding a new population:
\begin{equation}
\label{eq:pop_evolution}
\mathcal{P}_{t+1} = \text{ICSO}(\mathcal{P}_t, f(\mathcal{P}_t))
\end{equation}
which serves as the initialization for the next epoch. The optimal individual is then selected using the $f_{base}$, and its architecture parameters are used to update the super-net:
\begin{equation}
\label{eq:best_selection}
\bm{x}^* = \arg\min_{\bm{x} \in \mathcal{P}_{t+1}} f_{\text{base}}(\bm{x})
\end{equation}
\begin{equation}
\label{eq:arch_update}
\bm{\alpha}_{t+1} \leftarrow \bm{\alpha}_t + \eta_{\alpha} \cdot (\bm{x}^* - \bm{\alpha}_t)
\end{equation}

Following the architecture update, the supernet undergoes training to update the network weights:
\begin{equation}
\label{eq:weight_update}
w_{t+1} \leftarrow w_t - \eta_w \nabla_w \mathcal{L}_{\text{train}}(w_t, \bm{\alpha}_{t+1})
\end{equation}

The updated weights are then integrated with the evolved population to initiate the next iteration. This cyclic process continues until the validation accuracy exceeds a predefined threshold $T_{\text{stab}}$, at which point the framework transitions into the Stability Stage.

\subsection{Stable Stage}

In this subsection, we introduce the third phase: the Stability Stage. 

In this stage, the super-net reverts to the standard DARTS methodology, simultaneously updating both network weights $w$ and architecture parameters $\bm{\alpha}$. However, to facilitate fine-grained convergence towards a precise local optimum, a significantly reduced architecture learning rate is employed:
\begin{equation}
\label{eq:stability_update}
\bm{\alpha}_{t+1} \leftarrow \bm{\alpha}_t - \eta_{\alpha}^{\text{stab}} \nabla_{\bm{\alpha}} \mathcal{L}_{\text{val}}(w_t, \bm{\alpha}_t)
\end{equation}
where $\eta_{\alpha}^{\text{stab}}$ denotes the stability-phase architecture learning rate.

After stabilizing for a predetermined minimum number of epochs, the method initiates an architecture convergence verification based on Hoeffding's Inequality\cite{hoeffding1963probability}. Let $V_t = \|\bm{\alpha}_t - \bm{\alpha}_{t-1}\|_2$ denote the variation of architecture parameters at epoch $t$, and $\bar{V} = \frac{1}{n}\sum_{i=1}^{n} V_{t-n+i}$ be the mean variation within a sliding window of size $n$. The search terminates when:
\begin{equation}
\label{eq:hoeffding_stop}
\bar{V} < \epsilon = \sqrt{\frac{D \ln(2/\delta)}{2n}}
\end{equation}
where $D$ is the dimension of $\bm{\alpha}$, $\delta$ is the significance level, and $1-\delta$ is the confidence level.

\section{Experiment and Result Analysis}
In this section, we first assess the performance of G-ICSO-NAS within the DARTS search space using CIFAR-10, CIFAR-100, and ImageNet for image classification tasks. Subsequently, the effectiveness of the proposed method is further examined on the NAS-Bench-201 search space with CIFAR-10, CIFAR-100, and ImageNet-16-120 datasets.
\begin{table}[htbp]
    \centering
    \caption{Key Experiment Settings}
    \label{table1}
    \begin{tabular}{l l l}
        \toprule
        \textbf{Stage} & \textbf{Parameters} & \textbf{Value} \\
        \midrule
        \multirow{4}{*}{Warm up} 
            & batch size         & 64 \\
            & learning rate      & 0.025 \\
            & warm Up Epoch      & 5 \\
            & arch learning rate & 0 \\
        \midrule
        \multirow{5}{*}{Exploration} 
            & ICSO pop size      & 60 \\
            & ICSO $\phi$           & 0.15 \\
            & diversity weight   & 0.3 \\
            & operation diversity weight  & 0.2 \\
            & ICSO generations   & 8 \\
        \midrule
        \multirow{5}{*}{Stable} 
            & stability threshold  & 82\% \\
            & batch size         & 64 \\
            & stable arch learning rate & 1e-5 \\
            & early stop alpha threshold & 1e-3 \\
            & early stop confidence level & 0.95 \\
        \bottomrule
    \end{tabular}
\end{table}
\subsection{Experiment environment}
All experiments are carried out on a Dell Precision 3630 workstation equipped with an Intel Xeon E-2224 processor, 16~GB of system memory, and an NVIDIA RTX A4000 GPU with 16~GB VRAM. The experimental framework is implemented using Python~3.11.9 and PyTorch~2.7.1, with CUDA~12.8 support.

\begin{table*}[!t] 
\centering
\caption{Comparison of Neural Architecture Search Methods on CIFAR-10 and CIFAR-100}
\label{table2}
\begin{tabular}{lccccc}
\toprule
\multirow{2}{*}{Architecture/Method} & \multicolumn{2}{c}{Test Error (\%)} & Params & Search Cost & \multirow{2}{*}{Method Type} \\ 
\cmidrule(lr){2-3}
& CIFAR-10 & CIFAR-100 & (M) & (GPU-Days) & \\ 
\midrule
ResNet  & 4.61 & 22.1 & 1.7 & - & Manual \\
\midrule
ENAS + cutout  & 2.89 & - & 4.6 & 0.5 & RL \\
AmoebaNet-A & 3.34 & 18.93 & 3.2 & 3150 & Evolution \\
NSGA-Net  & 2.75 & 20.74 & 3.3 & 4.0 & Evolution \\
NSGANetV1-A2  & 2.65 & - & 0.9 & 27 & Evolution \\
EPCNAS-C  & 3.24 & 18.36 & 1.44 & 1.2 & Evolution \\
EAEPSO & 2.74 & 16.94 & 2.94 & 2.2 & Evolution \\
\midrule
DARTS (1st) & 3.00 & - & 3.3 & 1.5 & Gradient \\
DARTS (2st) & 2.76 & - & 3.3 & 4.0 & Gradient \\
SNAS (moderate) + cutout  & 2.85 & 17.55 & 2.8 & 1.5 & Gradient \\
ProxylessNAS + cutout  & \textbf{2.02} & - & - & 4.0 & Gradient \\
GDAS  & 2.93 & 18.38 & 3.4 & 0.2 & Gradient \\
BayesNAS  & 2.81 & - & 3.4 & 0.2 & Gradient \\
PC-DARTS + cutout  & 2.57 & 16.90 & 3.6 & \textbf{0.1} & Gradient \\
DARTS-  & 2.59 & 17.51 & 3.4 & 0.4 & Gradient \\
DrNAS  & 2.54 & - & 4.0 & 0.4 & Gradient \\
DARTS+PT  & 2.61 & - & 3.0 & 0.8 & Gradient \\
\midrule
G-ICSO-NAS & 2.54 & \textbf{16.90} & 4.24 & 0.15 & Hybrid \\
\bottomrule
\end{tabular}
\end{table*}

\begin{table}[htbp]
\centering
\caption{Comparison of Neural Architecture Search Methods on ImageNet}
\label{table3}
\begin{tabular}{lccc}
\toprule
\multirow{2}{*}{Architecture/Method} & Test Error & Test Error & Params \\
 & top-1(\%) & top-5(\%) & (M) \\
\midrule
ShuffleNet-V2 & 25.1 & 9.9 & 7.4 \\
MobileNet-V2 & 28.0 & 9.0 & 3.4 \\
\midrule
DARTS & 26.9 & 9.0 & 4.9 \\
DARTS(2st) & 26.7 & 8.7 & 4.7 \\
SNAS & 27.3 & 9.2 & 4.3 \\
GDAS & 26.0 & 8.5 & 5.3 \\
BayesNAS & 26.5 & 8.9 & 3.9 \\
PC-DARTS & 25.1 & 7.8 & 5.3 \\
SETN & 26.7 & 8.6 & 5.2 \\
NASNet-A & 26.0 & 8.4 & 5.3 \\
NASNet-B & 27.2 & 8.7 & 5.3 \\
NASNet-C & 27.5 & 9.0 & 4.9 \\
AmoebaNet-A & 25.5 & 8.0 & 5.1 \\
AmoebaNet-B & 26.0 & 8.5 & 5.3 \\
EoiNAS & 25.6 & 8.3 & 5.0 \\
MFENAS & 26.06 & 8.18 & 5.98 \\
SaDENAS & 25.08 & 7.98 & 5.58 \\
\midrule
G-ICSO-NAS & \textbf{24.98} & \textbf{7.49} & 5.8 \\
\bottomrule
\end{tabular}
\end{table}

\subsection{Datasets and Parameters}
The CIFAR-10 dataset \cite{krizhevsky2009learning}  contains  60{,}000 RGB images with a spatial resolution of $32 \times 32$, distributed across 10 mutually exclusive categories (e.g., airplane, automobile, bird, and cat). Each category includes 6{,}000 samples, which are divided into 50{,}000 images for training and 10{,}000 images for testing.

The CIFAR-100 dataset \cite{krizhevsky2009learning} adopts the same image resolution ($32 \times 32$) and overall sample size (60{,}000 images) as CIFAR-10, while posing a more challenging classification scenario with 100 distinct classes. Each class consists of 600 images, split into 500 training samples and 100 testing samples. The 100 fine-grained categories are further grouped into 20 coarse superclasses, making CIFAR-100 particularly suitable for both coarse- and fine-level recognition evaluation.

ImageNet \cite{deng2009imagenet}  is a large-scale hierarchical image dataset comprising 1{,}000 object categories and approximately 1.28 million training images, along with 50{,}000 validation samples and 100{,}000 test images. In contrast to CIFAR datasets, ImageNet images exhibit varying resolutions and are typically resized or center-cropped to a fixed spatial dimension (e.g., $224 \times 224$) prior to model input. As a result, ImageNet is widely regarded as a standard benchmark for evaluating large-scale visual recognition models.

Architecture search is conducted on CIFAR-10 under the DARTS search space, and the resulting architectures are subsequently evaluated across CIFAR-10, CIFAR-100, and ImageNet to assess their generalization performance. Detailed experimental settings for the search phase are listed in Table \ref{table1}.

NAS-Bench-201 \cite{dong2020bench} is adopted as a standardized evaluation benchmark to facilitate reproducible and computationally efficient comparisons among Neural Architecture Search (NAS) algorithms. Unlike benchmarks based on unbounded search spaces, NAS-Bench-201 employs a predefined and finite search space containing 15{,}625 candidate neural architectures. For each architecture, exhaustive training and evaluation results on CIFAR-10, CIFAR-100, and ImageNet-16-120 are provided. The benchmark further offers a query-based interface that enables direct access to ground-truth performance indicators, such as accuracy, training time, and parameter count, under different training budgets. This design removes the need for repeated architecture training during the search process, thereby substantially reducing computational overhead and mitigating randomness introduced by stochastic optimization. Following common practice, we conduct independent evaluations on NAS-Bench-201 using four distinct random seeds.

\begin{table*}[htbp]
\centering
\caption{Comparison of Neural Architecture Search Methods on NAS-Bench-201}
\label{table4}
\begin{tabular}{lcccccc}
\toprule
\multirow{2}{*}{Architecture/Method} & \multicolumn{2}{c}{CIFAR-10} & \multicolumn{2}{c}{CIFAR-100} & \multicolumn{2}{c}{ImageNet16-120} \\
\cmidrule(lr){2-3} \cmidrule(lr){4-5} \cmidrule(lr){6-7}
 & valid & test & valid & test & valid & test \\
\midrule
ResNet  & 90.83 & 93.97 & 70.42 & 70.86 & 44.53 & 43.63 \\
Random (baseline) & 90.93$\pm$0.36 & 93.70$\pm$0.36 & 70.60$\pm$1.37 & 70.65$\pm$1.38 & 42.92$\pm$2.00 & 42.96$\pm$2.15 \\
ENAS  & 37.51$\pm$3.19 & 53.89$\pm$0.58 & 13.37$\pm$2.35 & 13.96$\pm$2.33 & 15.06$\pm$1.95 & 14.57$\pm$2.10 \\
RandomNAS  & 80.42$\pm$3.58 & 84.07$\pm$3.61 & 52.12$\pm$5.55 & 52.31$\pm$5.77 & 27.22$\pm$3.24 & 26.28$\pm$3.09 \\
SETN  & 84.04$\pm$0.28 & 87.64$\pm$0.00 & 58.86$\pm$0.06 & 59.05$\pm$0.24 & 33.06$\pm$0.02 & 32.52$\pm$0.21 \\
GDAS  & 90.01$\pm$0.46 & 93.23$\pm$0.23 & 24.05$\pm$8.12 & 24.20$\pm$8.08 & 40.66$\pm$0.00 & 41.02$\pm$0.00 \\
DSNAS  & 89.66$\pm$0.29 & 93.08$\pm$0.13 & 30.87$\pm$16.40 & 31.01$\pm$16.38 & 40.61$\pm$0.09 & 41.07$\pm$0.09 \\
DARTS (1st)  & 39.77$\pm$0.00 & 54.30$\pm$0.00 & 15.03$\pm$0.00 & 15.61$\pm$0.00 & 16.43$\pm$0.00 & 16.32$\pm$0.00 \\
DARTS (2st)  & 39.77$\pm$0.00 & 54.30$\pm$0.00 & 15.03$\pm$0.00 & 15.61$\pm$0.00 & 16.43$\pm$0.00 & 16.32$\pm$0.00 \\
PC-DARTS  & 89.96$\pm$0.15 & 93.41$\pm$0.30 & 67.12$\pm$0.39 & 67.48$\pm$0.89 & 40.83$\pm$0.08 & 41.31$\pm$0.22 \\
iDARTS  & 89.86$\pm$0.60 & 93.58$\pm$0.32 & 70.57$\pm$0.24 & 70.83$\pm$0.48 & 40.38$\pm$0.59 & 40.89$\pm$0.68 \\
DARTS- & \textbf{91.03$\pm$0.44}& 93.80$\pm$0.40 & 71.36$\pm$1.51 & 71.53$\pm$1.51 & 44.87$\pm$1.46 & 45.12$\pm$0.82 \\
\midrule
G-ICSO-NAS & 90.22$\pm$0.12 & \textbf{94.00$\pm$0.08} & \textbf{72.17$\pm$0.38} & \textbf{72.63$\pm$0.47} & \textbf{45.21$\pm$0.26} & \textbf{45.66$\pm$0.34} \\
\midrule
Optimal & 91.61 & 94.37 & 73.49 & 73.51 & 46.77 & 47.31 \\
\bottomrule
\end{tabular}
\end{table*}

\subsection{Experiment Result and Comparison on DARTS search space}
The comparative performance of G-ICSO-NAS against representative NAS approaches on CIFAR-10 and CIFAR-100 is summarized in Table~\ref{table2}.

On CIFAR-10, G-ICSO-NAS achieves a test error of 2.54\% while requiring only 0.15 GPU-days for architecture search, demonstrating clear advantages over established baselines such as DARTS baseline, SNAS, BayesNAS, and NSGA-Net. Although ProxylessNAS \cite{cai2018proxylessnas} report marginally lower error rates, their search costs are substantially higher, reaching 4.0, respectively, which exceeds that of G-ICSO-NAS by a notable margin.

To further examine the transferability of the architectures discovered on CIFAR-10, the selected model is evaluated on CIFAR-100 and ImageNet. As summarized in Table~\ref{table2}, G-ICSO-NAS achieves a test error of 16.90\% on CIFAR-100, exhibiting performance comparable to PC-DARTS while outperforming several representative NAS baselines.

Moreover, the corresponding evaluation results on ImageNet are reported in Table~\ref{table3}, where G-ICSO-NAS demonstrates competitive performance in terms of both top-1 and top-5 error rates.

\subsection{Experiment Result and Comparison on NAS-Bench-201 search space}

The comparative evaluation of G-ICSO-NAS against representative state-of-the-art NAS approaches on the NAS-Bench-201 benchmark is reported in Table~\ref{table4}, covering experiments on CIFAR-10, CIFAR-100, and ImageNet-16-120.

Across the NAS-Bench-201 search space, G-ICSO-NAS exhibits strong overall performance. The architecture identified based on CIFAR-10 achieves the best results on six evaluation metrics, outperforming a wide range of existing NAS methods. 

Overall, G-ICSO-NAS demonstrates consistently favorable performance across extensive evaluations conducted on both the DARTS and NAS-Bench-201 search spaces, as well as on CIFAR-10, CIFAR-100, and ImageNet datasets. When combined with its ultra-low search cost of only 0.15 GPU-days in the DARTS search space, these results indicate that G-ICSO-NAS effectively achieves a balanced trade-off between search efficiency and architecture quality, aligning with the design objectives of the proposed framework.

\subsection{Ablation study}
To further analyze the contribution of individual components within the proposed G-ICSO-NAS framework, ablation experiments are conducted.

Regarding the parameter diversity fitness and operation diversity fitness, extensive experiments demonstrate that setting the weights of both diversity functions to zero leads to considerable instability in the search results. A significant proportion of such cases exhibit an excessive presence of none and skip-connect operations, which adversely affect the final architecture's training performance.

About the Stability Stage and Early Stopping Strategy, experimental results indicate that after 10-15 epochs of ICSO-based search (with 8 generations per epoch), the explored optimal architecture tends to stabilize. Continuing ICSO iterations beyond this point substantially increases computational cost without meaningful improvement. Furthermore, employing a reduced architecture learning rate during the stability stage helps prevent the re-emergence of dominant none and skip-connect operations.

The warm-up stage is also essential. Multiple experiments show that omitting the warm-up stage or using only 1--2 epochs results in severe fluctuations in architecture performance during the exploration stage, causing the validation accuracy to remain below the stability stage threshold for an extended period. This occurs because the super-net weights $w$ undergo substantial updates during the initial epochs. If the exploration stage begins prematurely, the architecture parameters $\alpha$ also experience drastic updates, which disrupts the optimization direction of the ICSO algorithm.

\section{Conclusion}
In this work, we present G-ICSO-NAS, a neural architecture search framework that jointly incorporates evolutionary optimization and gradient-based learning. The proposed approach adopts a three-stage search paradigm, consisting of a warm-up phase that exclusively optimizes supernet weights while fixing architecture parameters to establish a stable initialization, an exploration phase that integrates ICSO-based architecture optimization with gradient descent through an alternating update scheme, and a stabilization phase that performs fine-grained architecture refinement assisted by Hoeffding-based early stopping to enable reliable convergence detection.

Comprehensive evaluations on both the DARTS and NAS-Bench-201 search spaces, spanning CIFAR-10, CIFAR-100, and ImageNet datasets, consistently demonstrate that G-ICSO-NAS achieves a favorable trade-off between search efficiency and architectural performance. Notably, with an ultra-low search cost of only 0.15 GPU-days under the DARTS search space, the proposed method attains competitive or superior results compared with state-of-the-art NAS approaches.

Future research will focus on further improving the robustness and performance of the proposed framework.Additionally, we intend to extend G-ICSO-NAS to other search tasks beyond image classification.

\FloatBarrier
\bibliographystyle{IEEEtran}
\bibliography{ref}

\end{document}